\documentclass{article}

\usepackage{arxiv}

\usepackage[utf8]{inputenc} 
\usepackage[T1]{fontenc}    
\usepackage{hyperref}       
\usepackage{url}            
\usepackage{booktabs}       
\usepackage{amsfonts}       
\usepackage{nicefrac}       
\usepackage{microtype}      
\usepackage{lipsum}		
\usepackage{graphicx}
\usepackage{natbib}
\usepackage{doi}

\title{OVLW-DETR: Open-Vocabulary Light-Weighted Detection Transformer}

\author{
Yu Wang, Xiangbo Su, Qiang Chen, Xinyu Zhang, Teng Xi, \\ 
\textbf{Kun Yao, Errui Ding, Gang Zhang, Jingdong Wang} \\
Baidu Inc., China \\
\texttt{\{wangyu106,suxiangbo, chenqiang13, zhangxinyu14,}\\
\texttt{xiteng01, yaokun01, dingerrui, zhanggang03, wangjingdong\}@baidu.com}
}

\renewcommand{\headeright}{Technical Report}
\renewcommand{\undertitle}{Technical Report}
\renewcommand{\shorttitle}{OVLW-DETR: Open-Vocabulary Light-Weighted Detection Transformer}

\hypersetup{
pdftitle={OVLW-DETR: Open-Vocabulary Light-Weighted Detection Transformer},
pdfsubject={computer vision},
pdfauthor={Yu Wang},
pdfkeywords={Open Vocabulary Object Detection},
}

\begin{document}
\maketitle

\begin{abstract}
Open-vocabulary object detection focusing on detecting novel categories guided by natural language. In this report, we propose Open-Vocabulary Light-Weighted Detection Transformer (OVLW-DETR), a deployment friendly open-vocabulary detector with strong performance and low latency. Building upon OVLW-DETR, we provide an end-to-end training recipe that transferring knowledge from vision-language model (VLM) to object detector with simple alignment. We align detector with the text encoder from VLM by replacing the fixed classification layer weights in detector with the class-name embeddings extracted from the text encoder. Without additional fusing module, OVLW-DETR is flexible and deployment friendly, making it easier to implement and modulate. improving the efficiency of interleaved attention computation. Experimental results demonstrate that the proposed approach is superior over existing real-time open-vocabulary detectors on standard Zero-Shot LVIS benchmark. Source code and pre-trained models are available at [https://github.com/Atten4Vis/LW-DETR].
\end{abstract}


\section{Introduction}
Traditional object detection methods are typically trained on a set of predefined categories, which limits the application in real-world scenarios with open vocabulary. With the development of vision-language model with contrastive image-text training, many recent works\cite{zhang2022glipv2}\cite{Yao_2023_CVPR}\cite{minderer2024scaling} aim to transfer the language capabilities of VLM models\cite{radford2021learning} to object detection. However, the computation deployment consumption continue to pose challenges for the practical application of ov-detector in real-world scenarios.
The key of open-vocabulary detection is on transferring the language knowledge in VLM to object detector. In recent works, \cite{gu2021open} distills the VLM-predictions on image regions, \cite{liu2023grounding}\cite{zhang2022glipv2} introduce deep fusion module between the image and language encoders. However, these methods suffers from high inference lantency. \cite{cheng2024yolo} propose the first real-time open-vocabulary based on Yolov8\cite{Jocher_Ultralytics_YOLO_2023}, the text-image fusion module still results in a reduction of the inference speed.

In this report, we introduce Open-Vocabulary Light-Weighted Detection Transformer (OVLW-DETR), a DETR-based real-time OV detector. This work is built upon the recent LW-DETR\cite{chen2024lw}, a light-weight real-time DETR baseline. We validate the effectiveness of generalizing real-time DETR to open-vocabulary task with simple alignment, achieving real-time inference speed and strong open-vocabulary performance. In OVLW-DETR, the language knowledge in VLM is transferred to DETR by replacing the classification layer weights in detector with the class-name embeddings extracted from the text encoder. Without additional fusion module, the model retains its inherent architecture, thereby enhancing its flexibility and facilitating a more efficient deployment.

Our results show that OVLW-DETR achieves a strong performance and low latency on open-vocabulary task with a simple alignment. It outperforms prior work on both open-vocabulary and long-tail detection benchmarks. On open-vocabulary LVIS benchmark, OVLW-DETR obtains.

\section{OVLW-DETR}
\subsection{Architecture}
\begin{figure}
	\centering
	\includegraphics[scale=0.45]{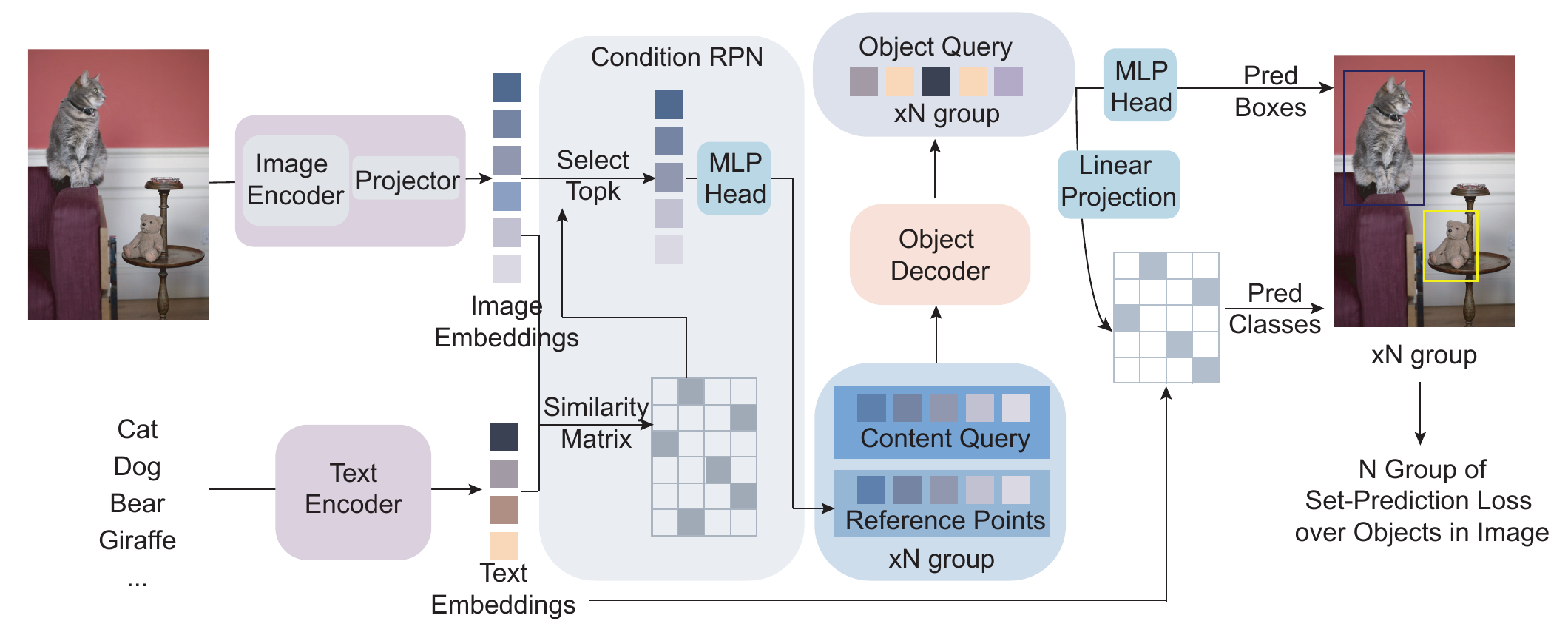}
	\caption{OVLW-DETR framework}
	\label{fig:fig1}
\end{figure}

The overall framework of OVLW-DETR is presented in Figure \ref{fig:fig1}, which consists of a LW-DETR detector and a text encoder. The detector retrains the inherent LW-DETR architecture, consisting of a ViT encoder, a projector, and a DETR decoder. The text encoder is adopted from the text branch of pretrained VLM model. Open-vocabulary classification is enabled by simple alignment between detector and VLM text encoder, modifying the the mixed-query selection scheme and classification layer. The fixed layer weights in both selecting top-K spatial queries scheme and query classification layers are replaced  with the text embedding extracted by the text encoder. 
\subsection{Training Scheme}
We fine-tune the pre-trained detector on standard LW-DETR training receipt, while the text encoder is frozen during training process to retain the generalization ability. For stable and efficient training, we adopt the IoU-aware classification loss, IA-BCE loss and parallel weight-sharing decoders, Group DETR in training scheme.

\section{Experiment}
\subsection{Implementation Details}
The detector is initialized with pre-trained LW-DETR, and the text encoder is initialized with pre-trained CLIP. We propose a series of S/M/L OVLW-DETR based on LW-DETR S/M/L, and we adopt the text encoder from CLIP ConvNeXt-Large\cite{Ilharco_OpenCLIP_2021}.

In training process, we adopt the default LW-DETR training schedule, with AdamW optimizer with an initial learning rate of 1e-4 and weight decay of 1e-4. We fine-tune the whole detector for 12 epochs with total batch size of 64, while freeze the text encoder for generalization.

Training dataset consists of GoldG\cite{kamath2021mdetr}(including Flickr30K \cite{plummer2015flickr30k} and GQA\cite{hudson2019gqa}) and Object365\cite{shao2019objects365} . We mix GoldG and VG randomly at a ratio of 80\% to 20\% in our experiments. Specifically, we randomly sample $10\%$ O365 every epoch. During training process, we use the default augmentation of DETR, consisting of random flip, random crop, and resizing the input to $640\times 640$ resolution.
\subsection{Evaluation}
We use LVIS v1.0 minival\cite{kamath2021mdetr} as benchmark since this dataset contains 1203 categories, with a long tail of rare categories. We evaluate the standard AP with 300 evaluation queries. We measure the averaged inference latency in an end-to-end manner with fp16 precision and a batch size of 1 on LVIS v1 minival on one T4 GPU, with 1203 categories. The performance and the end-to-end latency are measured for all real-time detectors with official deploying implementations.
\subsection{Results}
The main results are illustrated in Table \ref{tab:table}. We mainly compare OVLW-DETR with YOLO-World\cite{cheng2024yolo}, the recent state-of-the art real-time open-vocabulary method. OVLW-DETR achieves or outperforms on zero-shot LVIS minival benchmark on different scales. Without multi-modal fusion module, OVLW-DETR maintains the low latency in inference while also demonstrating impressive performance.
\begin{table}[ht]
	\caption{\textbf{Zero-Shot Evaluation on LVIS minival}. We report standard AP on LVIS minival. The latency is tested on T4 GPU with FP16 precision on LVIS categories. We test the latency of YOLO-Worldv2 without NMS.}
	\centering
	\begin{tabular}{ccccccccc}
     
		\toprule
		\cmidrule(r){1-2}
		Model     & Backbone  & Data & AP & APr & APc & APf & Params & Latency\\
		\midrule
		GLIPv2 & Swin-T & O365,GoldG,Cap4M & 29.0  & - & - & - & 232M&-  \\
            \midrule
	    YOLO-World-S & YOLOv8-S & O365,GoldG & 24.3	& 16.6 & 22.1 & 27.7 & 13M & -\\
            YOLO-World-M & YOLOv8-M & O365,GoldG & 28.6 & 19.7 & 26.6 & 31.9 & 29M & -\\
            YOLO-World-L & YOLOv8-L & O365,GoldG & 32.5 & 22.3 & 30.6 & 36.1 & 48M & -\\
            YOLO-World-L & YOLOv8-L & O365,GoldG,CC3M-Lite & 33.0 & 23.6 & 32.0 & 35.5 & 48M & -\\
            \midrule
            YOLO-Worldv2-S & YOLOv8-S & O365,GoldG & 22.7 & 16.3 & 20.8 & 25.5 & 13M & 5.58\\
            YOLO-Worldv2-M & YOLOv8-M & O365,GoldG & 30.0 & \textbf{25.0} & 27.2 & \textbf{33.4} & 29M & 9.54\\
            YOLO-Worldv2-L & YOLOv8-L & O365,GoldG & 33.0 & 22.6 & 32.0 & \textbf{35.8} & 48M & 13.66\\
            YOLO-Worldv2-L & YOLOv8-L & O365,GoldG,CC3M-Lite & 32.9 & 25.3 & 31.1 & 35.8 & 48M & 13.66\\
            \midrule
            OVLW-DETR-S & LW-DETR-S & O365,GoldG & \textbf{27.3} & \textbf{22.8} & \textbf{26.7} & \textbf{28.6} & 12M & 3.23\\
            OVLW-DETR-M & LW-DETR-M & O365,GoldG & \textbf{30.2} & 24.0 & \textbf{28.9} & 32.4 & 28M & 6.09\\
            OVLW-DETR-L & LW-DETR-L & O365,GoldG & \textbf{33.5} & \textbf{26.5} & \textbf{33.9} & 34.4 & 47M & 9.90\\
		\bottomrule
	\end{tabular}
	\label{tab:table}
\end{table}

\section{Conclusion}
In this paper, we propose OVLW-DETR, an open-vocabulary light-weighted DETR detector. With simple alignment, our method validate the generalization of transferring real-time detection transformers to open-vocabulary task. We hope our report can provide insights for building open-vocabulary detector.

\bibliographystyle{unsrtnat}
\bibliography{template}  






\end{document}